\documentclass{article}
\usepackage{spconf,amsmath,epsfig}
\usepackage{hyperref}
\usepackage[utf8]{inputenc}
\usepackage{tikz}
\usepackage{booktabs}
\usepackage{tabularx}
\usepackage{subcaption}
\usepackage{mathtools}
\usepackage[capitalise]{cleveref}
\usepackage{enumitem}

\def\x{{\mathbf x}}

\newcolumntype{Y}{>{\centering\arraybackslash}X}

\title{HOW USEFUL IS REGION-BASED CLASSIFICATION OF REMOTE SENSING IMAGES IN A DEEP LEARNING FRAMEWORK ?}
\name{Nicolas Audebert\textsuperscript{1, 2}, Bertrand Le Saux \textsuperscript{1}, Sébastien Lefèvre \textsuperscript{2}}
\address{\textsuperscript{1 }ONERA, \textit{The French Aerospace Lab}, F-91761 Palaiseau, France\\\textsuperscript{2 }Univ. Bretagne-Sud, UMR 6074, IRISA, F-56000 Vannes, France\\\small{Emails: \{nicolas.audebert, bertrand.le\_saux\}@onera.fr, sebastien.lefevre@irisa.fr}
}
	
\begin{document}

\maketitle
\begin{abstract}
In this paper, we investigate the impact of segmentation algorithms as a preprocessing step for classification of remote sensing images in a deep learning framework. Especially, we address the issue of segmenting the image into regions to be classified using pre-trained deep neural networks as feature extractors for an SVM-based classifier. An efficient segmentation as a preprocessing step helps learning by adding a spatially-coherent structure to the data. Therefore, we compare algorithms producing superpixels with more traditional remote sensing segmentation algorithms and measure the variation in terms of classification accuracy. We establish that superpixel algorithms allow for a better classification accuracy as a homogenous and compact segmentation favors better generalization of the training samples.
\end{abstract}


\section{INTRODUCTION}
\label{sec:intro}
Last years have seen the rise of deep learning approaches for computer vision and remote sensing is not an exception. However, deep networks are not designed to directly process high resolution images such as the ones used in remote sensing. These models are therefore used by focusing on the local appearance around a given location. For performance reasons, a preprocessing step dividing the image into coherent small regions is needed, and therefore arise the need for segmentation. Two main approaches are used in the litterature~: sliding windows and image segmentation. Superpixel segmentations gained lots of interest in the context of remote sensing when it was used to establish state-of-the-art performances (both in classification accuracy and processing time) \cite{paisitkriangkrai_effective_2015-igarss}.

In this paper, we investigate what makes a good segmentation algorithm for classification purposes. Two aspects are evaluated: the pure segmentation quality (well-defined boundaries, coherence of the pixels inside a region) and the impact of the segmentation on classification through the size and shape of the regions (i.e the classification samples).

\section{CLASSIFICATION FRAMEWORK}
\label{sec:classif}

\begin{figure*}[t]
  \includegraphics[width=\textwidth]{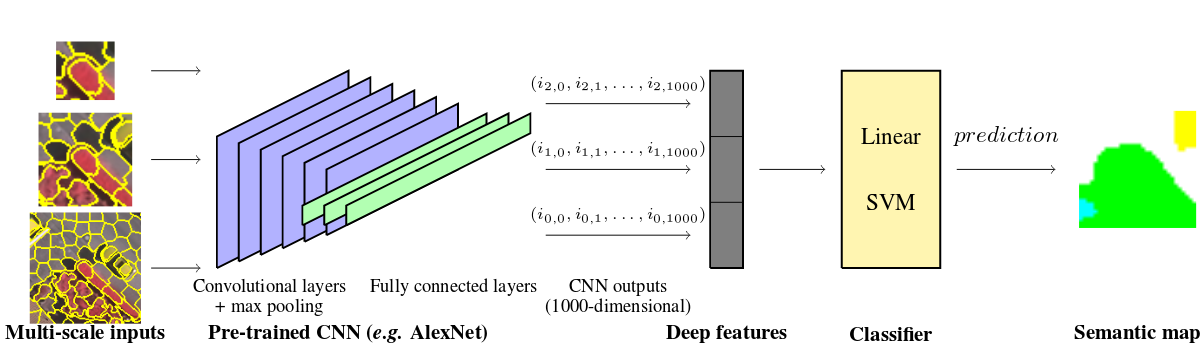}
  \caption{Classification framework using deep multi-scale features}
  \label{fig:framework}
\end{figure*}

Deep learning has been the state-of-the-art in computer vision for a few years. Neural networks operate at the pixel level by simultaneously learning which features to extract and how to classify them. Convolutional neural networks are composed of layers of neurons computing convolutions on the previous layer outputs. These layers are stacked and combined with max-pooling (i.e. sampling maximum activations) and elementwise non-linear transfer functions (e.g. $tanh$) to extract high order features from the input. The network then produces a probability vector, on which a softmax is applied to predict the output label. This has been proven to be an effective baseline for computer vision tasks in \cite{razavian_cnn_2014-baseline-igarss}.

Our framework uses the well-known AlexNet \cite{krizhevsky_imagenet_2012-igarss} architecture as a feature extractor, as \cite{lagrange_benchmarking_2015-igarss} showed that the deep features extracted from AlexNet could be effectively transfered for remote sensing tasks. Patches extracted from the image are passed through the network and the last layer outputs before the softmax are used as feature vectors. More precisely, our framework (\cref{fig:framework}) achieves semantic segmentation with the following pipeline :
  \begin{enumerate}[noitemsep,nolistsep]
    \item Divide the image into small regions using a segmentation algorithm.
    \item For each region, extract $32 \times 32$, $64 \times 64$ and $128 \times 128$ patches centered on the region.
    \item Resize all patches to $228 \times 228$ and process them through AlexNet.
    \item Concatenate the resulting vectors to produce one feature vector (sample).
  \end{enumerate}
  
At training time, we process the images of the training set, for which we have the associated ground truth. We define the label of a region with a majority vote according to the associated ground truth. We then use the training set of newly acquired features to train a linear Support Vector Machine (SVM), whose parameters are optimized by stochastic gradient descent. At testing time, we use the SVM to predict the label of each region of the image to be classified, and then associate to all pixels in this region the predicted output label. In the end, we obtain a semantic map that we can compare to the ground truth.

\section{SEGMENTATION ALGORITHMS}
\label{sec:seg}
To avoid discrepancies in the training samples, the segmented regions should be similar in shape and size. This motivates the use of superpixel algorithms rather than traditional segmentation ones. Indeed, the latter, both from the remote sensing and computer vision communities, create very inhomogenous regions in shape and size. This does not bode well with our multiscale framework, that expects similarly shaped training samples. Moreover, superpixel algorithms have been used successfully in the remote sensing literature \cite{wu_superpixel-based_2012-igarss}. Therefore, we choose to evaluate the following superpixel algorithms:
\begin{itemize}[noitemsep,nolistsep]
  \item \textbf{SLIC} (Simple Linear Iterative Clustering) \cite{achanta_slic_2010-igarss}: starts from a grid and creates a segmentation by iteratively growing the regions by applying a $k$-means algorithm.
  \item \textbf{LSC} (Linear Spectral Clustering) \cite{li_superpixel_2015-igarss}: embeds the image in a 14-dimension space and increase each region using weighted $k$-means starting from a grid.
  \item Quickshift \cite{vedaldi_quick_2008-igarss}: clusters points belonging to the same dominant mode in a non-Euclidean color-(x,y) space, using the Lab color space.
\end{itemize}
We also test two popular segmentation algorithms from the remote sensing community:
\begin{itemize}[noitemsep,nolistsep]
  \item \textbf{MRS} (Multiresolution Segmentation) \cite{baatz_multiresolution_2000-igarss}: implemented in the eCognition software, MRS clusters points using a well-defined homogeneity criterion based on spatial and spectral information.
  \item \textbf{HSEG} (Hierarchical image Segmentation) \cite{tilton_best_2012-igarss}: based on Hierarchical Step-Wise Optimization (HSWO) with spectral clustering, HSEG builds a hierarchical segmentation using a dissimilarity criterion. We extract the most detailed segmentation using the RHSEG implementation.
\end{itemize}
As a baseline, we compare these segmentations to a sliding window (SW) approach. The window parameters are chosen to obtain as many windows as there are regions using the previously described algorithms to achieve the same processing time.

\section{EXPERIMENTS}
\label{sec:res}

\begin{figure*}[t]
  \centering
  \begin{subfigure}{0.13\textwidth}
    \hfill
  \end{subfigure}
  \begin{subfigure}{0.13\textwidth}
    \hfill
  \end{subfigure}
  \begin{subfigure}{0.13\textwidth}
    \includegraphics[width=\textwidth]{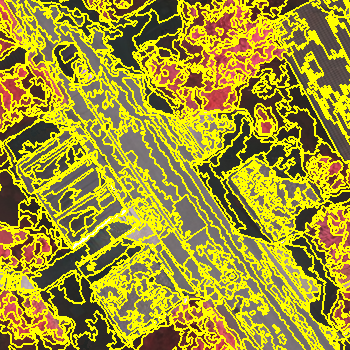}
    \caption{MRS}
  \end{subfigure}
  \begin{subfigure}{0.13\textwidth}
    \includegraphics[width=\textwidth]{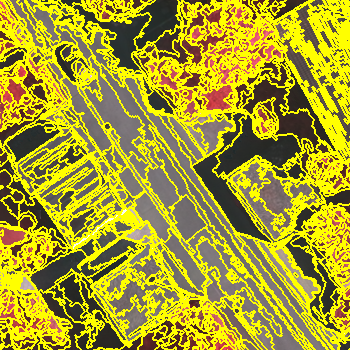}
    \caption{HSEG}
  \end{subfigure}
  \begin{subfigure}{0.13\textwidth}
    \includegraphics[width=\textwidth]{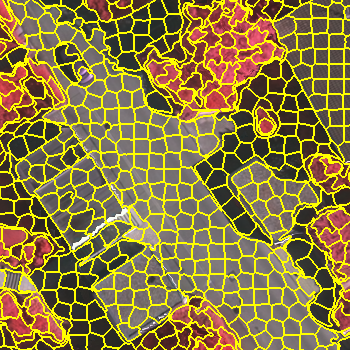}
    \caption{SLIC}
  \end{subfigure}
  \begin{subfigure}{0.13\textwidth}
    \includegraphics[width=\textwidth]{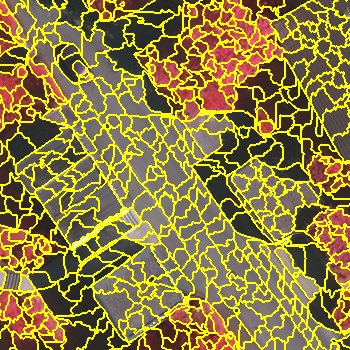}
    \caption{Quickshift}
  \end{subfigure}
  \begin{subfigure}{0.13\textwidth}
    \includegraphics[width=\textwidth]{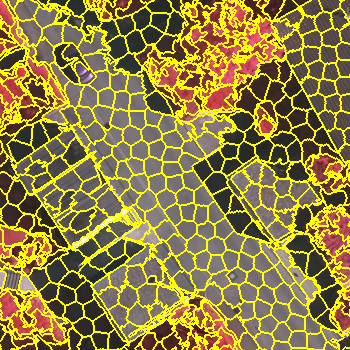}
    \caption{LSC}
  \end{subfigure}
  \caption{Regions segmented by the different segmentation algorithms (zoom on a specific location)}
  \label{fig:segs}
\end{figure*}

\begin{figure*}[t]
  \centering
  \begin{subfigure}{0.13\textwidth}
    \includegraphics[width=\textwidth]{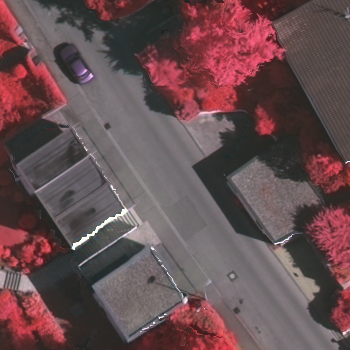}
    \caption{Orthoimage}
  \end{subfigure}
  \begin{subfigure}{0.13\textwidth}
    \includegraphics[width=\textwidth]{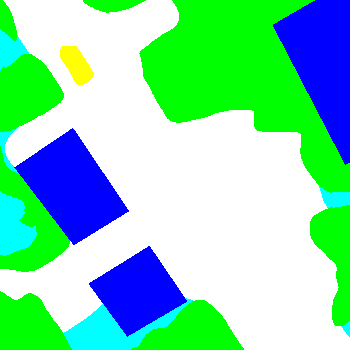}
    \caption{Ground truth}
  \end{subfigure}
  \begin{subfigure}{0.13\textwidth}
    \includegraphics[width=\textwidth]{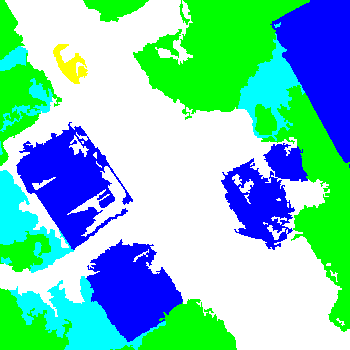}
    \caption{MRS}
  \end{subfigure}
  \begin{subfigure}{0.13\textwidth}
    \includegraphics[width=\textwidth]{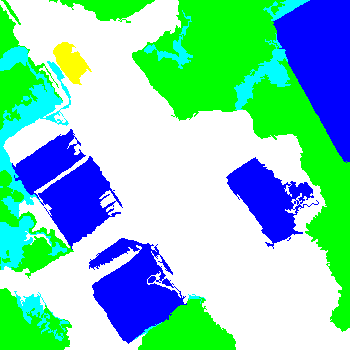}
    \caption{HSEG}
  \end{subfigure}
  \begin{subfigure}{0.13\textwidth}
    \includegraphics[width=\textwidth]{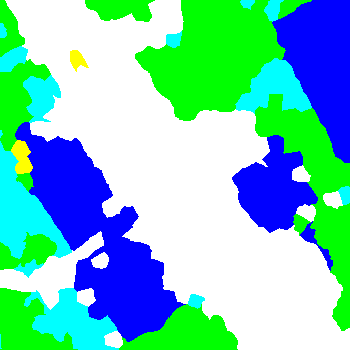}
    \caption{SLIC}
  \end{subfigure}
  \begin{subfigure}{0.13\textwidth}
    \includegraphics[width=\textwidth]{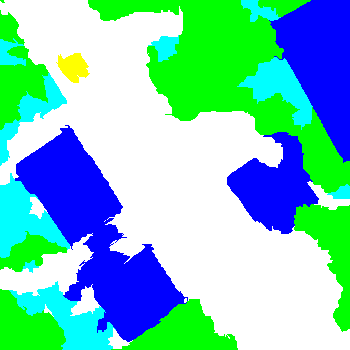}
    \caption{Quickshift}
  \end{subfigure}
  \begin{subfigure}{0.13\textwidth}
    \includegraphics[width=\textwidth]{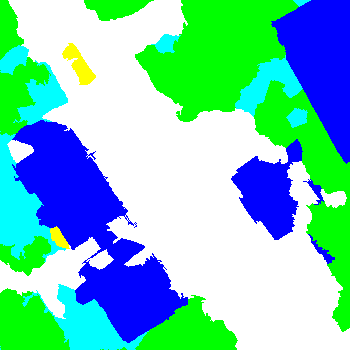}
    \caption{LSC}
  \end{subfigure}
  \caption{Semantic maps after classification using different segmentation algorithms (zoom on a specific location)}
  \label{fig:patches}
\end{figure*}

\subsection{Experimental setup}

The algorithms are tested on the ISPRS 2D Semantic Labeling Dataset \cite{rottensteiner_isprs_2012-igarss}. We use part of the Vaihingen data, consisting of 16 IR-R-G orthoimages with pixel-level ground truth. We compare the segmented images to the ideal segmentation represented by the ground truth.

Segmentation algorithms are evaluted by several standard metrics proposed by \cite{neubert_superpixel_2012-igarss}:
\begin{itemize}[noitemsep,nolistsep]
  \item The Undersegmentation Error (\textbf{UE}): defined as the ratio of pixels belonging to a region overlapping other regions. Formally, if respectively $S$, $P$ and $N$ denote the regions in the ground truth, the segmented regions and the number of pixels in the image:
  \[ UE = \frac{1}{N} \sum_{S \in GT} \sum_{P : P \cap S \neq \emptyset} min(|P \cap S|, |P \backslash P \cap S|)\]
   \item The Boundary Recall (\textbf{BR}): the recall of boundary pixels in the 3-pixel neighborhood of the ground truth boundaries :
  \[ BR = \frac{true\ pos.}{true\ pos. + false\ neg.} \]
  \item The Average Purity (\textbf{AP}): average percentage of pixels of a region belonging to the region dominant class. Let $avg$ and $maj$ denote respectively the average function and the majority class:
  \[ AP = \underset{P \in seg}{avg} (\frac{| P \cap maj(P)|}{| P |}) \]
  \item The oracle: the pixel-wise classification accuracy that would be achieved by a perfect classifier, assigning the majority class label to each segment. This is the best case scenario and therefore is the maximum accuracy that can be achieved with this segmentation.
\end{itemize}

We split this dataset as follow : tiles 1, 5, 7, 11, 17, 23, 26, 28, 34 and 37 form the training set, while tiles 13, 21 and 30 form the validation set and tiles 3, 15 and 32 form the testing set. Note that the ``clutter" class is not represented in the testing set. This is justified by the fact that the ISPRS evaluation procedure does not take this class into account.

In order to compare the classification results, we use the following metrics:
\begin{itemize}[noitemsep,nolistsep]
  \item The overall pixel-wise accuracy on the testing set.
  \item The $\kappa$ coefficient (inter-rater agreement).
  \item The F1 score for the ``car" class, as an additionnal performance indicator. This allows us to consider specifically the problem of object detection.
\end{itemize}
Note that the segmentation parameters for each algorithm were chosen to achieve the best overall accuracy as we wish to optimize the classification performance, using roughly the same number of regions.

\subsection{Results}

\Crefname{table}{Tab.}{Tabs.}

The segmentation metrics of the evaluated algorithms are presented in \Cref{table:seg_metrics}. Using only this information, MRS and HSEG seem to be competitive with superpixel algorithms on most metrics.

The classification results for each algorithm are presented in \Cref{table:classif_metrics}. The superpixel algorithms obtain very similar results as the classification accuracy shows little variation w.r.t. the segmentation. However, these results establish an advantage of superpixels over traditional segmentations. Indeed, MRS and HSEG are lagging behind the superpixel methods on the classification accuracy and do not bring any additional gain compared to the baseline sliding window approach. This can be explained by the segmentation's geometrical properties. Superpixels tend be strongly convex and compact, while traditional segmentations usually produce very heterogenous regions in shape and size. However, better learning is achieved when the training samples are similar, as the classifier does not need to infer which pixels are meaningful in the example patch. The parameters achieving best classification accuracy for MRS back this result. To reach the accuracy presented in \Cref{table:classif_metrics}, the compacity parameter of MRS has to be significantly increased and the resulting segmentation is more homogenous and ``superpixel-looking". However, MRS needs significantly more segments than the superpixel algorithms (especially efficient ones such as SLIC) -- which comes at the cost of a higher processing time -- while the accuracy is still lower than with superpixel algorithms.

This is partially illustrated in \cref{fig:segs} and \cref{fig:patches}. MRS and HSEG are more erratic and the semantic maps suffer from irregular shapes and borders. The interior of objects such as cars and buildings is often attributed to the wrong class since the inside pixels do not belong to the same regions as the sucessfully classified ones. Superpixel algorithms tend to preserve more truthfully the shape and convexity of objects.

Furthermore, there is no direct link between the theoretical best-case (the oracle) and the actual accuracy. LSC has a lower oracle than SLIC on the dataset, but beats it in the at testing time. This means that the choice of the algorithm not only impacts the segmentation but also the information learned by the classifier. Indeed, the shape and size of the superpixels directly alter the samples provided to the SVM. This supports the idea that the homogeneity of the superpixels is crucial in this classification framework.

Finally, according to the resulting F1 scores on the ``car" pixels, object detection can greatly be improved by chosing an appropriate segmentation algorithm. Best results on this class are obtained with LSC, even if results are tight.

\begin{table}
  \centering
  \setlength{\tabcolsep}{3pt}
  \begin{tabularx}{0.5\textwidth}{ Y c c c c c }
  \toprule
  Algorithm & Regions & UE (\%) & BR (\%) & AP (\%) & Oracle (\%)\\
  \midrule
  SLIC & \textbf{20 000} & \textbf{10.21} & 84.07 & 75.10 & 89.91\\
  LSC & 22 800 & 11.37 & 91.13 & 71.54 & 85.83\\
  Quickshift & 21 000 & 11.66 & 88.34 & 72.90 & 83.61\\
  \cmidrule(lr){1-6}
  MRS & 23 500 & 13.12 & \textbf{95.71} & \textbf{79.08} & \textbf{91.68}\\
  HSEG & 21 000 & 11.39 & 94.83 & 78.66 & 85.25\\
  \bottomrule
  \end{tabularx}
  \caption{Segmentation metrics on the ISPRS dataset}
  \label{table:seg_metrics}
  \vspace{0.75em}
  
  \begin{tabularx}{0.5\textwidth}{ Y Y Y Y Y }
  \toprule
  Algorithm & Regions & Acc. (\%) & F1\_car & $\kappa$\\  \midrule
  SLIC & \textbf{20 000} & 82.20 & 0.54 & 0.76\\
  LSC & 22 800 & \textbf{82.45} & \textbf{0.58} & 0.76\\
  Quickshift & 21 000 & 82.05 & 0.52 & 0.75\\
  \cmidrule(lr){1-5}
  MRS & 23 500 & 80.53 & 0.56 & 0.73\\
  HSEG & 21 000 & 79.56 & 0.54 & 0.72\\
  \cmidrule(lr){1-5}
  SW & 23 800 & 81.22 & 0.53 & 0.74\\
  \bottomrule
  \end{tabularx}
  \caption{Classification metrics on the ISPRS dataset}
  \label{table:classif_metrics}
  \vspace*{-1.25em}
\end{table}

\section{CONCLUSION}
\label{sec:conclusion}
In this work, we have aimed to establish that superpixel algorithms provide adequate segmentations for classification of remote sensing images in a deep learning framework. There is no clear universal advantage of using one particular superpixel segmentation method. This depends on the nature of the data, notably if distinguishing objects significantly smaller than others, such as cars compared to buildings, is needed.

This comparison brings new insights on how samples should be extracted from remote sensing data in order to achieve semantic segmentation, i.e segmentation and classification of the regions through a deep learning framework. Superpixel algorithms provide the classifier with compact and homogeneously segmented samples that favors generalization of the learned content. This allows for a better accuracy with fewer samples and a reduced processing time.

Finally, our work shows that there is no direct link between the quality of the segmentation according to the standard metrics (boundary adherence, etc.) and the pixel-wise classification accuracy. Therefore, chosing a segmentation algorithm should be based solely on the classification accuracy achieved, as the impact of the shape, size and homogeneity of the segments is preponderant for training a classifier.

\section{ACKNOWLEDGEMENTS}
\label{sec:ack}

The Vaihingen dataset was provided by the German Society for Photogrammetry, Remote Sensing and Geoinformation (DGPF) (\url{http://www2.isprs.org/commissions/comm3/wg4/semantic-labeling.html}). Nicolas Audebert's work is supported by the Total-ONERA project NAOMI.


\bibliographystyle{IEEEbib}
\bibliography{refs-igarss}

\end{document}